\documentclass[11pt,a4paper]{article}
\usepackage[hyperref]{acl2019}
\usepackage{times}
\usepackage{latexsym}
\usepackage{amsmath}
\usepackage{verbatim}
\usepackage{enumitem}
\usepackage{float}
\usepackage{graphicx}
\usepackage{makecell}
\usepackage{url}
\usepackage{capt-of}

\usepackage[font=small,labelfont=bf]{caption}

\aclfinalcopy 

\newcommand{\bx}{\mathbf{x}}
\newcommand{\bh}{\mathbf{h}}

\newcommand{\bs}{\mathbf{s}}

\newcommand{\br}{\mathbf{r}}

\newcommand{\relenc}{f_{\theta}}

\newcommand{\bert}{\textsc{bert}}
\newcommand{\bertem}{\textsc{bert}$_{\textsc{EM}}$}
\newcommand{\mtbmodel}{\textsc{bert}$_{\textsc{EM}}$\textsc{+mtb}}
\newcommand{\blank}{[\textsc{blank}]}
\newcommand{\cls}{[\textsc{cls}]}
\newcommand{\sep}{[\textsc{sep}]}

\title{Matching the Blanks: Distributional Similarity for Relation Learning}

\author{Livio Baldini Soares \qquad Nicholas FitzGerald \qquad Jeffrey Ling\thanks{\hspace{0.1em} Work done as part of the Google AI residency.} \qquad Tom Kwiatkowski \vspace{0.1cm} \\  Google Research \vspace{0.1cm} \\  \texttt{\{liviobs,nfitz,jeffreyling,tomkwiat\}@google.com}}
\date{}

\begin{document}
\maketitle

\begin{abstract}

General purpose relation extractors, which can model arbitrary relations, are a core aspiration in information extraction.
Efforts have been made to build general purpose extractors that represent relations with their surface forms, or which jointly embed surface forms with relations from an existing knowledge graph. 
However, both of these approaches are limited in their ability to generalize.
In this paper, we build on extensions of Harris' distributional hypothesis to relations, as well as recent advances in learning text representations (specifically, BERT), to build task agnostic relation representations solely from entity-linked text.
We show that these representations significantly outperform previous work on exemplar based relation extraction (FewRel) even without using any of that task's training data.
We also show that models initialized with our task agnostic representations, and then tuned on supervised relation extraction datasets, significantly outperform the previous methods on SemEval 2010 Task 8, KBP37, and TACRED.

\end{abstract}

\section{Introduction}
\label{sec:intro}
Reading text to identify and extract relations between entities has been a long standing goal in natural language processing \cite{cardie1997}.
Typically efforts in relation extraction fall into one of three groups. 
In a first group, supervised \cite{kambhatla2004combining,guodong2005exploring,C14-1220}, or distantly supervised relation extractors \cite{,mintz2009distant} learn a mapping from text to relations in a limited schema.
Forming a second group, open information extraction removes the limitations of a predefined schema by instead representing relations using their surface forms \cite{banko2007, fader2011, stanovsky2018}, which increases scope but also leads to an associated lack of generality since many surface forms can express the same relation. 
Finally, the universal schema \cite{riedel2013relation} embraces both the diversity of text, and the concise nature of schematic relations, to build a joint representation that has been extended to arbitrary textual input \cite{toutanova2015}, and arbitrary entity pairs \cite{verga2016}. 
However, like distantly supervised relation extractors, universal schema rely on large knowledge graphs (typically Freebase~\cite{bollacker2008}) that can be aligned to text.

Building on \citet{lin2001}'s extension of Harris' distributional
hypothesis~\cite{harris1954distributional} to relations, as well as recent advances in learning word representations from observations of their contexts \cite{mikolov2013, peters2018deep, devlin2018bert}, we propose a new method of learning relation representations directly from text. 
First, we study the ability of the Transformer neural network architecture \cite{vaswani2017attention} to encode relations between entity pairs, and we identify a method of representation that outperforms previous work in supervised relation extraction.
Then, we present a method of training this relation representation without any supervision from a knowledge graph or human annotators by \emph{matching the blanks}.

\begin{figure}[H]
    \centering
    \footnotesize
    \begin{tabular}{p{0.9\linewidth}}
     \blank{}, inspired by Cale's earlier cover, recorded one of the most acclaimed versions of ``\blank{}'' \\ \\
     \blank{}'s rendition of ``\blank{}'' has been called ``one of the great songs'' by Time, and is included on Rolling Stone's list of ``The 500 Greatest Songs of All Time''.
    \end{tabular}
    \caption{``Matching the blanks'' example where both relation statements share the same two entities.}
    \label{tab:pithy-example}
\end{figure}

Following \citet{riedel2013relation}, we assume access to a corpus of text in which entities have been linked to unique identifiers and we define a \emph{relation statement} to be a block of text containing two marked entities.
From this, we create training data that contains relation statements in which the entities have been replaced with a special \blank{} symbol, as illustrated in Figure~\ref{tab:pithy-example}.
Our training procedure takes in pairs of blank-containing relation statements, and has an objective that encourages relation representations to be similar if they range over the same pairs of entities.
After training, we employ learned relation representations to the recently released FewRel task \cite{han2018fewrel} in which specific relations, such as {\it `original language of work'} are represented with a few exemplars, such as {\it The Crowd (Italian: La Folla) is a 1951 Italian film}. 
\citet{han2018fewrel} presented FewRel as a supervised dataset, intended to evaluate models' ability to adapt to relations from new domains at test time. 
We show that through training by matching the blanks, we can outperform \citet{han2018fewrel}'s top performance on FewRel, without having seen any of the FewRel training data.
We also show that a model pre-trained by matching the blanks and tuned on FewRel outperforms humans on the FewRel evaluation. 
Similarly, by training by matching the blanks and then tuning on labeled data, we significantly improve performance on the SemEval 2010 Task 8 \cite{hendrickx2009semeval}, KBP-37 \cite{ZhangW15a}, and TACRED \cite{zhang2017position} relation extraction benchmarks.

\newcommand{\entstart}[1]{\langle}
\newcommand{\entend}[1]{\textsc{[\textbackslash e#1]}}

\section{Overview}\label{sec:overview}
\paragraph{Task definition}
In this paper, we focus on learning mappings from \emph{relation statements} to \emph{relation representations}. Formally, let $\bx = [x_0\dots x_n]$ be a sequence of tokens, where $x_0 = \cls$ and $x_n = \sep$ are special start and end markers. 
Let $\bs_1 = (i,j)$ and $\bs_2 = (k, l)$ be pairs of integers such that $0< i < j-1$, $j<k$, $k\leq l -1$, and $l \leq n$.
A relation statement is a triple $\br = (\bx, \bs_1, \bs_2)$, where the indices in $\bs_1$ and $\bs_2$ delimit entity mentions in $\bx$: the sequence $[x_i\dots x_{j-1}]$ mentions an entity, and so does the sequence $[x_k\dots x_{l-1}]$.
Our goal is to learn a function 
$\bh_r = \relenc(\br)$ that maps the relation statement to a fixed-length vector $\bh_{r} \in \mathcal{R}^d$ that represents the relation expressed in $\bx$ between the entities marked by $\bs_1$ and $\bs_2$.

\paragraph{Contributions}
This paper contains two main contributions.
First, in Section~\ref{sec:relation-tasks} we investigate different architectures for the relation encoder $\relenc$, all built on top of the widely used
Transformer sequence model~\cite{devlin2018bert,vaswani2017attention}.
We evaluate each of these architectures by applying them to a suite of relation extraction benchmarks with supervised training.

Our second, more significant, contribution---presented in Section~\ref{sec:mtb}---is to show that $\relenc$ can be learned from widely available distant supervision in the form of entity linked text.

\begin{figure*}[p]
  \centering
  \newcolumntype{P}[1]{>{\centering\arraybackslash}p{#1}}
  \begin{tabular}{P{0.45\textwidth}  P{0.45\textwidth}}
    \includegraphics[width=0.28\textwidth]{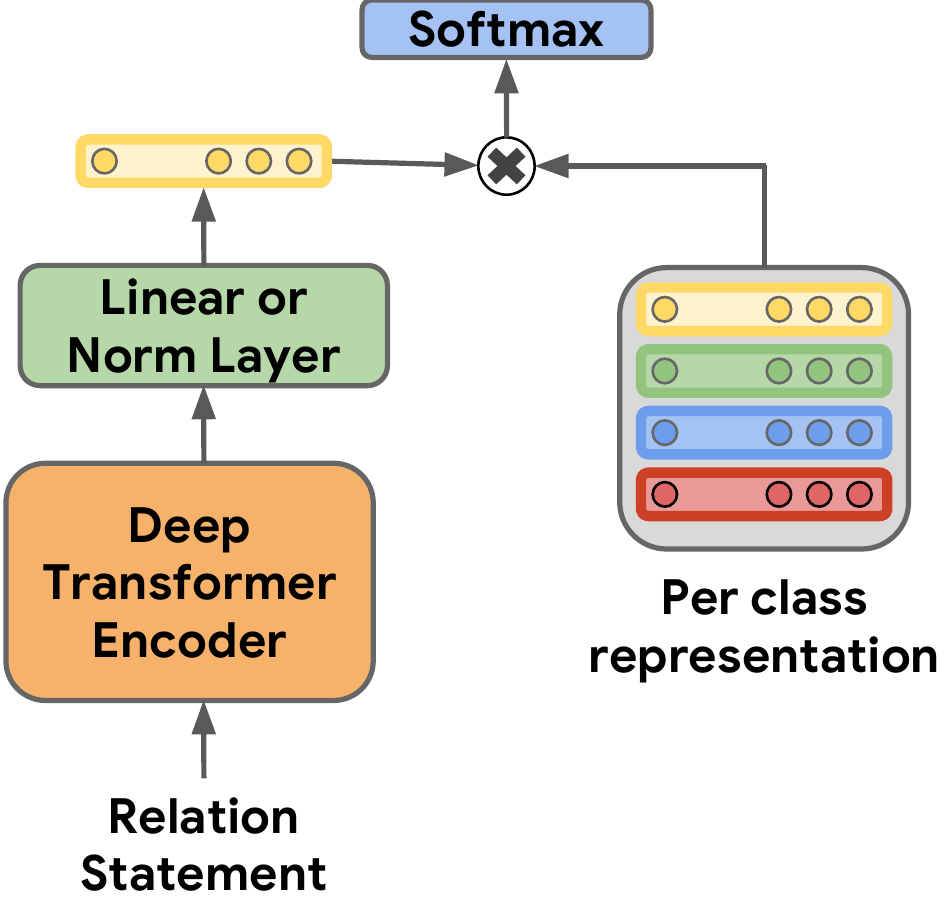} &
    \includegraphics[width=0.28\textwidth]{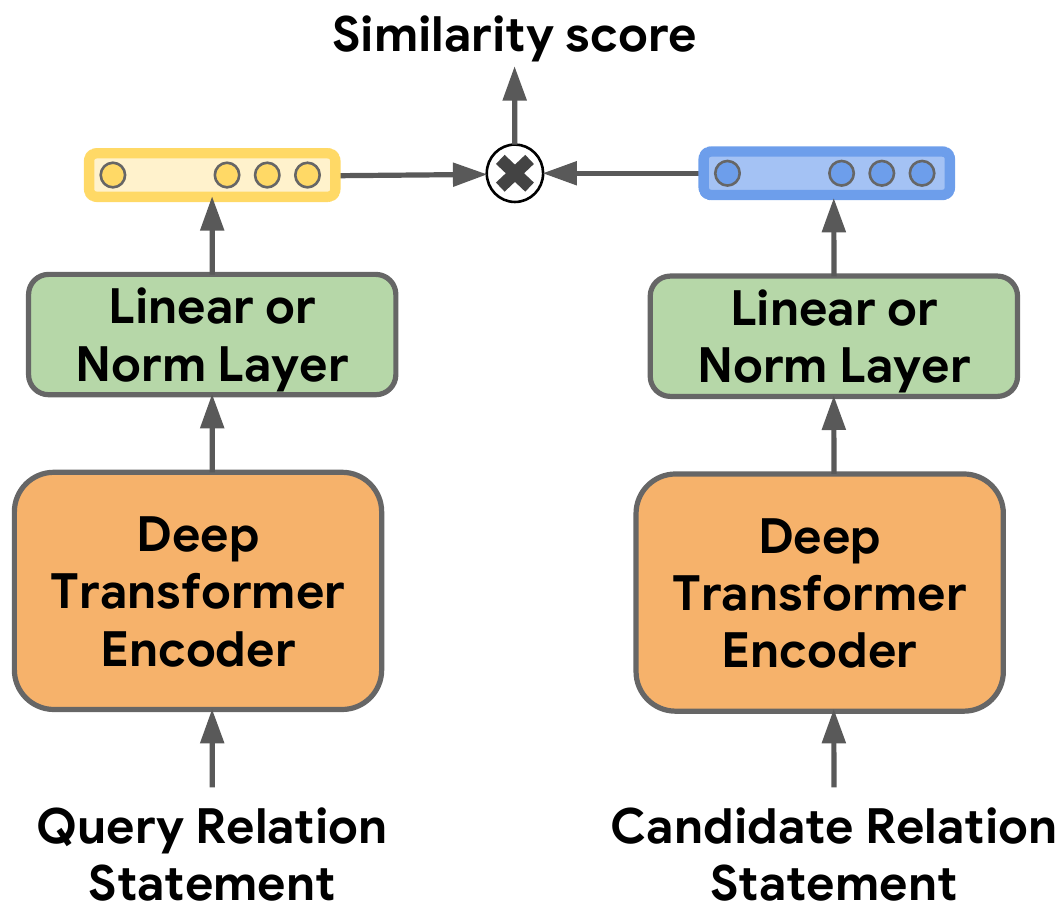} \\ \hline
  \end{tabular}
  \caption{Illustration of losses used in our models. The left figure depicts
  a model suitable for supervised training, where the model is expected to classify
  over a predefined dictionary of relation types. The figure on the right depicts a pairwise
  similarity loss used for few-shot classification task.}
  \label{fig:loss-architecture}
\end{figure*}

\begin{figure*}[p]
  \centering
  \footnotesize
  \begin{tabular}{ccc}
    \includegraphics[width=0.25\textwidth]{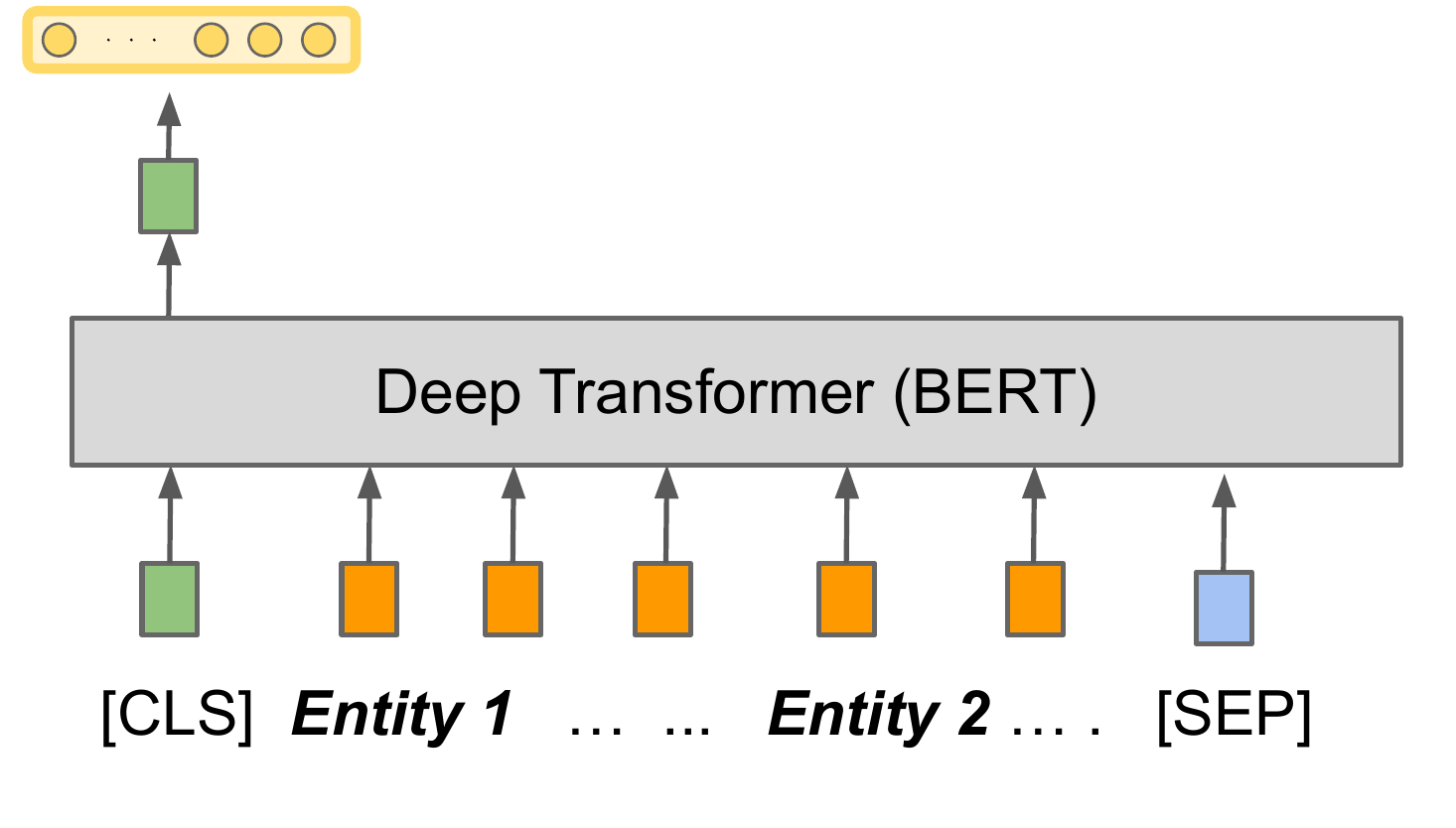} &
    \includegraphics[width=0.25\textwidth]{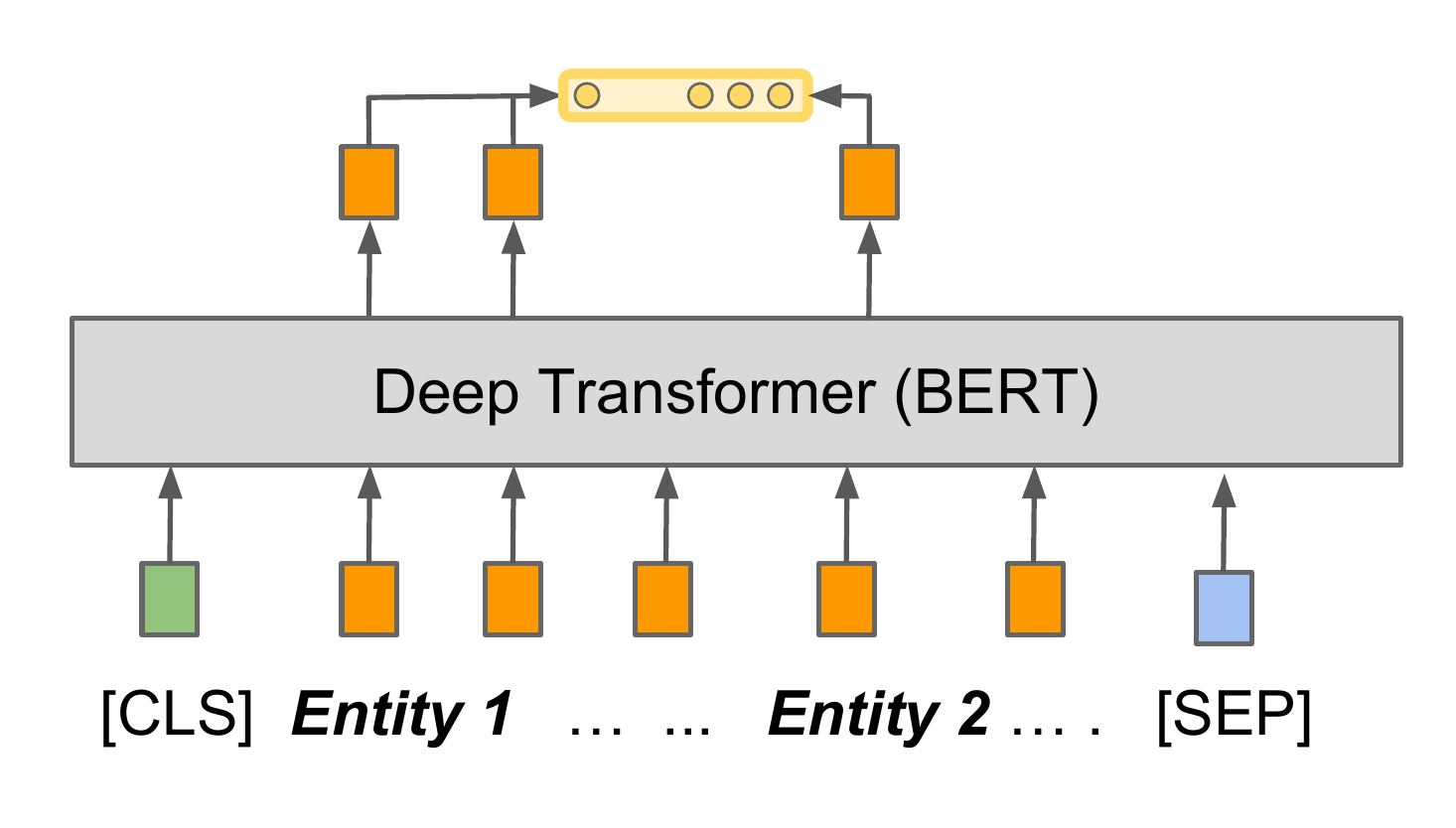} &
    \includegraphics[width=0.25\textwidth]{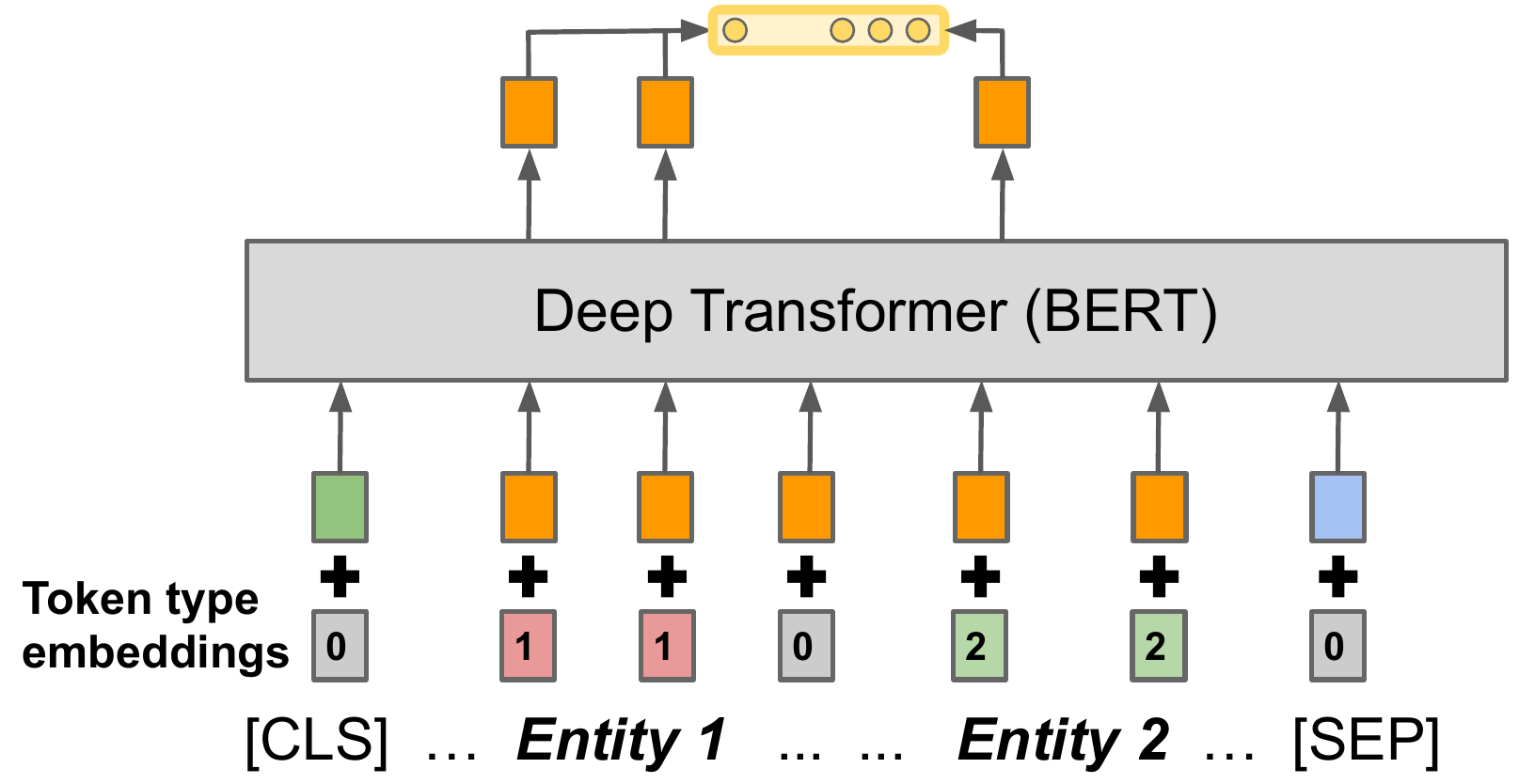} \\
    (a) \textsc{standard} -- \textsc{[cls]} & 
    (b) \textsc{standard} -- \textsc{mention pooling} &
    (c) \textsc{positional emb.} -- \textsc{mention pool.} \\ [10pt] 
    \includegraphics[width=0.25\textwidth]{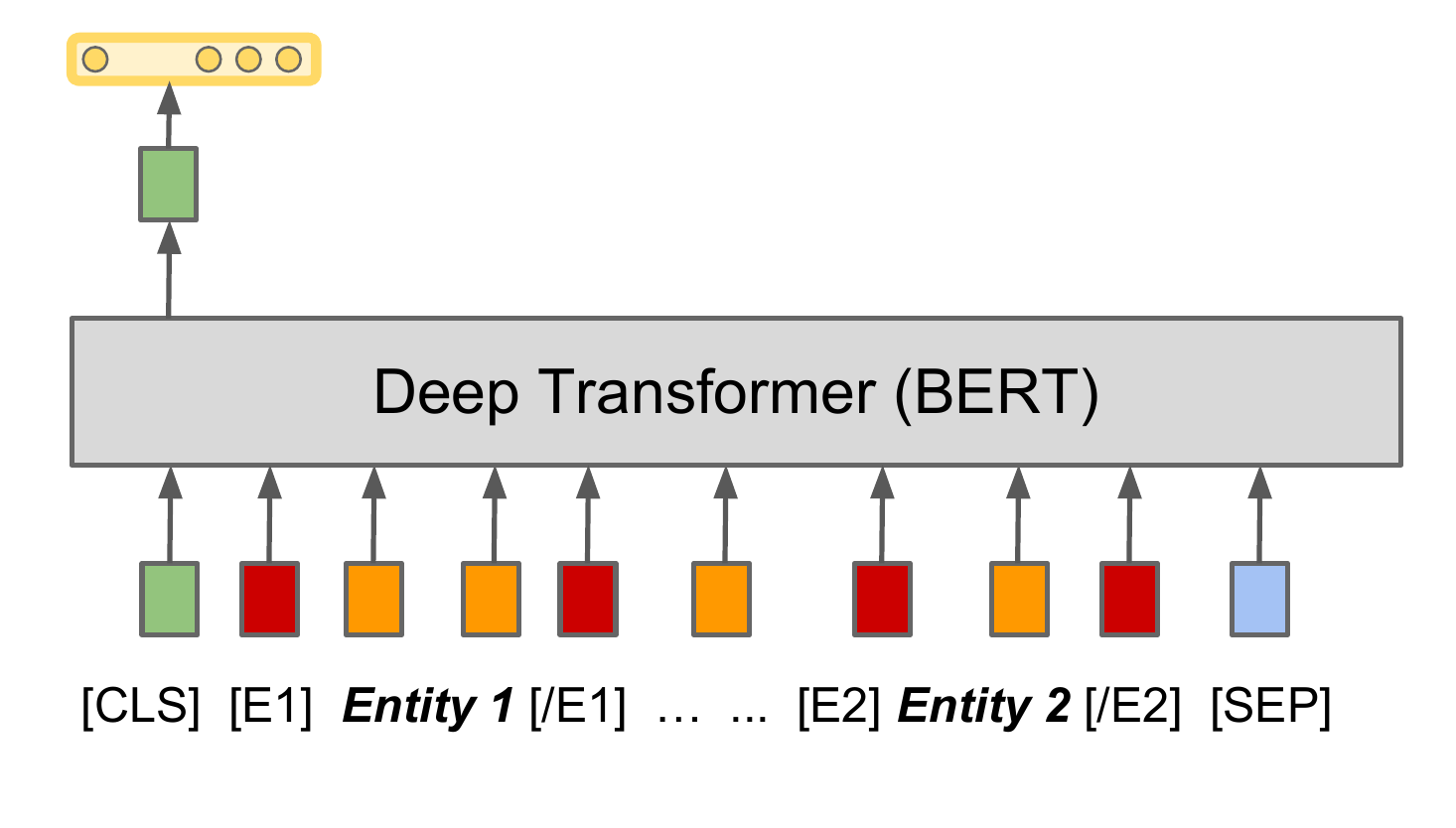} &
    \includegraphics[width=0.25\textwidth]{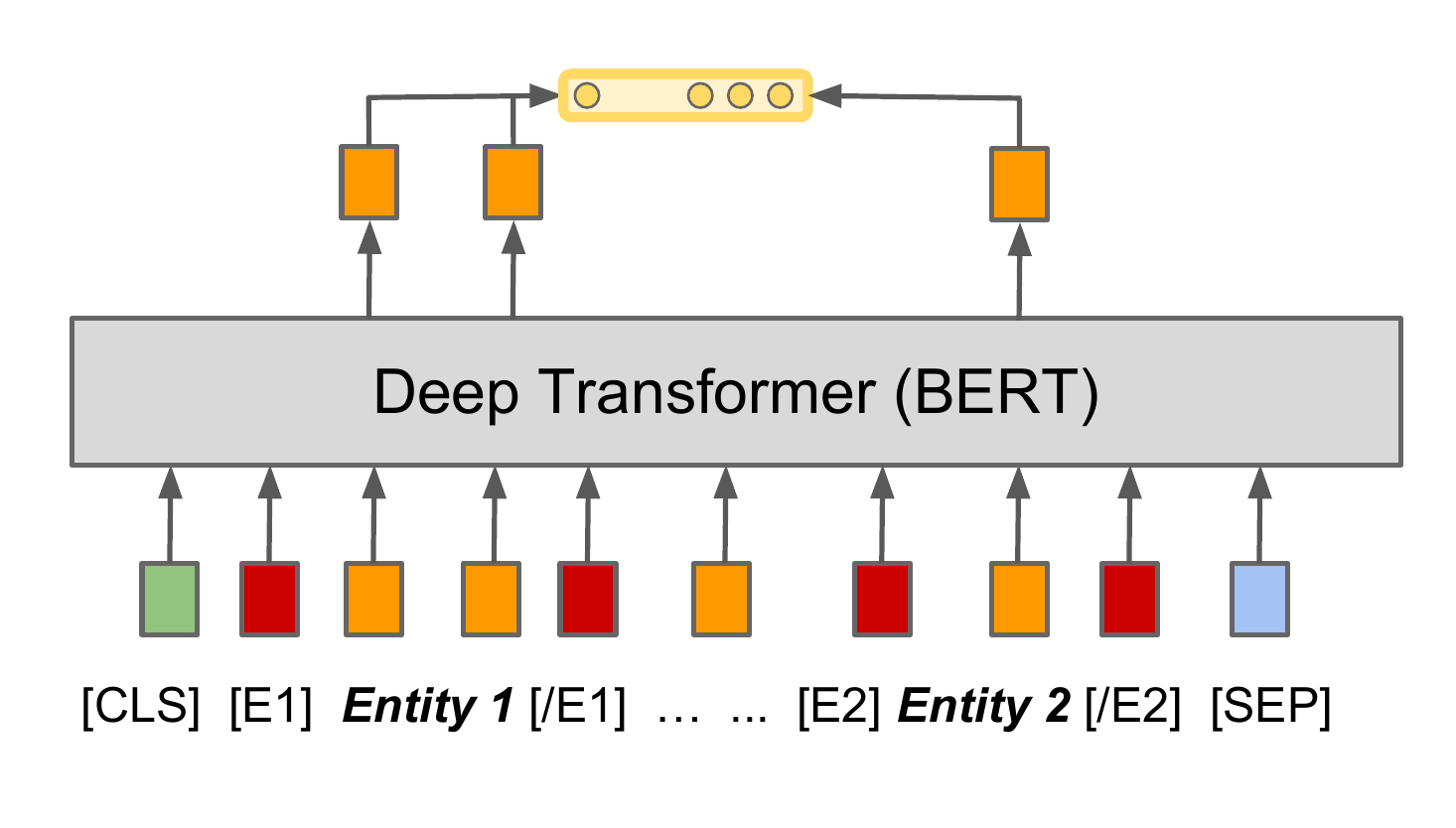} &
    \includegraphics[width=0.25\textwidth]{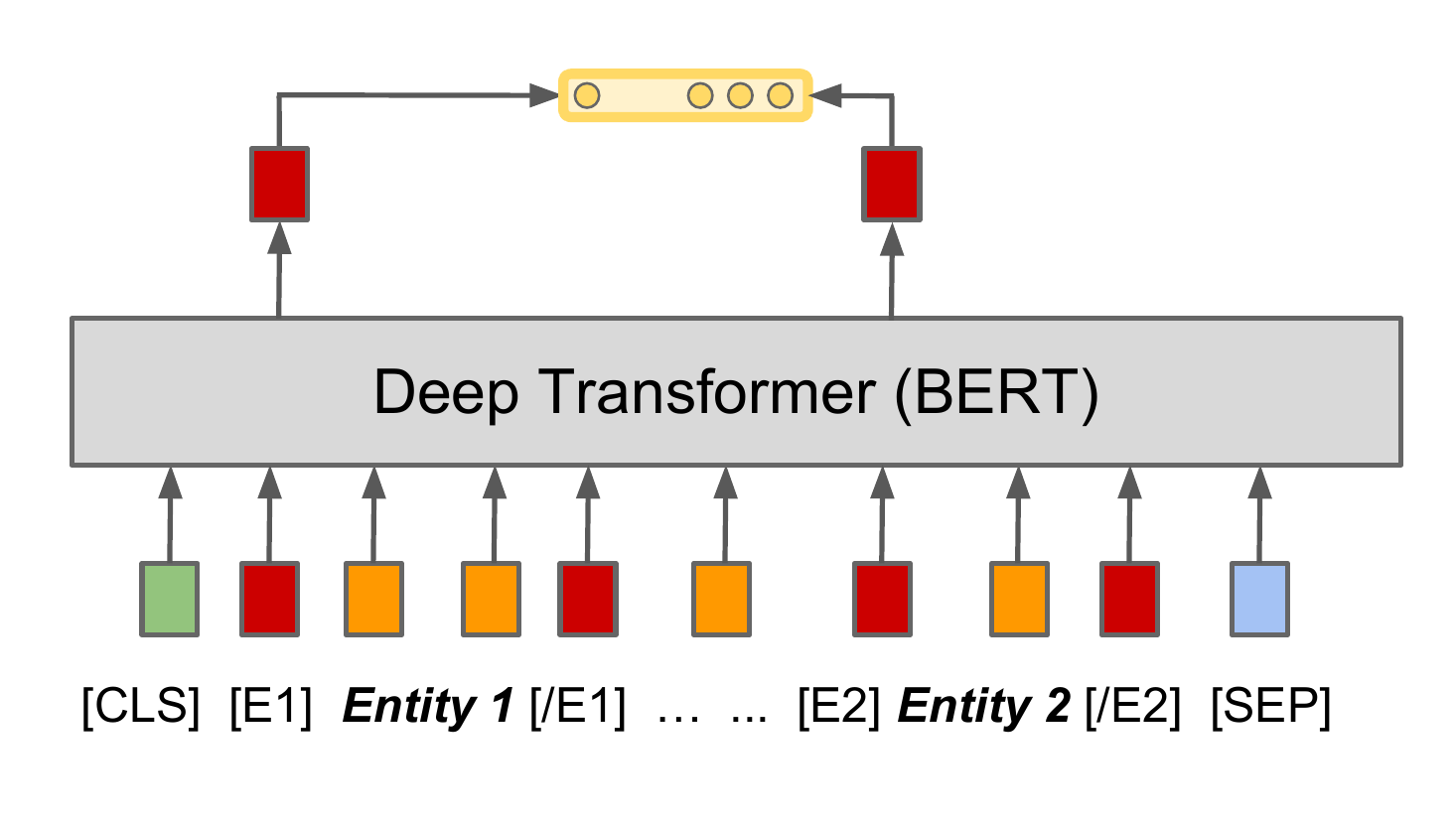} \\
    (d) \textsc{entity markers} -- \textsc{[cls]} &
    (e) \textsc{entity markers} -- \textsc{mention pool.} &
    (f) \textsc{entity markers} -- \textsc{entity start} \\ \hline
  \end{tabular}
  \caption{Variants of architectures for extracting relation representations from deep
  Transformers network. Figure~(a) depicts a model with \textsc{STANDARD} input and 
  \textsc{[CLS]} output, Figure~(b) depicts a model with \textsc{STANDARD} input and 
  \textsc{mention pooling} output and Figure~(c) depicts a model with
  \textsc{positional embeddings} input and \textsc{mention pooling} output.
  Figures (d), (e), and (f)  use \textsc{entity markers} input while using
  \textsc{[CLS]}, \textsc{mention pooling}, and \textsc{entity start} output, respectively.}
  \label{fig:model-variants}
\end{figure*}

\begin{table*}[p]
    \setlength{\tabcolsep}{5pt}
    \centering
    \footnotesize
    \begin{tabular}{|c|c|cc|cc|cc|c|}
    \hline
    \multicolumn{2}{|c|}{}
        & \multicolumn{2}{c|}{\textbf{SemEval 2010}}
        & \multicolumn{2}{c|}{\textbf{KBP37}}
        & \multicolumn{2}{c|}{\textbf{TACRED}} 
        & \textbf{FewRel} \\
    \multicolumn{2}{|c|}{}
        & \multicolumn{2}{c|}{\textbf{Task 8}}
        & \multicolumn{2}{c|}{}
        & \multicolumn{2}{c|}{}
        & \textbf{5-way-1-shot} \\        \hline
\multicolumn{2}{|c|}{\textbf{\# training annotated examples}} &
 \multicolumn{2}{c|}{8,000 (6,500 for dev)}    & 
 \multicolumn{2}{c|}{15,916}  & 
 \multicolumn{2}{c|}{68,120}  &
 44,800\\ \hline
\multicolumn{2}{|c|}{\textbf{\# relation types}} &
 \multicolumn{2}{c|}{19}  & 
 \multicolumn{2}{c|}{37}  & 
 \multicolumn{2}{c|}{42}  &
 100 \\ \hline
\multicolumn{2}{|c|}{} & Dev F1 & Test F1 & Dev F1 & Test F1 & Dev F1 & Test F1 & Dev Acc. \\ \hline
\multicolumn{2}{|c|}{\citet{P16-1123}*}  & -- & 88.0 & -- & -- & -- & -- & -- \\ \hline 
\multicolumn{2}{|c|}{\citet{ZhangW15a}*} & -- & 79.6 & -- & 58.8 & -- & -- & -- \\ \hline
\multicolumn{2}{|c|}{\citet{DBLP:journals/corr/abs-1807-03052}*} & -- & 84.8 & -- & -- & -- & 68.2 & -- \\ \hline
\multicolumn{2}{|c|}{\citet{han2018fewrel}}       & --    & -- & -- & -- & -- & -- & 71.6 \\ \hline \hline
\textbf{Input type}  & \textbf{Output type}       &        &    &       &     &       &   & \\ \hline
\textsc{standard} & \textsc{[CLS]}                & 71.6  & -- & 41.3  & --  & 23.4  & --  & 85.2 \\ \hline
\textsc{standard} & \textsc{mention pool.}        & 78.8  & -- & 48.3  & --  & 66.7  & --  & 87.5 \\ \hline
\textsc{positional emb.} & \textsc{mention pool.} & 79.1  & -- & 32.5  & --  & 63.9  & --  & 87.5 \\ \hline
\textsc{entity markers} & \textsc{[CLS]}          & 81.2  & --	& 68.7  & --  & 65.7  & --  & 85.2 \\ \hline
\textsc{entity markers} & \textsc{mention pool.}  & 80.4  & -- & 68.2  & --  & 69.5  & --  & 87.6 \\ \hline
\textsc{entity markers} & \textsc{entity start} & \textbf{82.1} & \textbf{89.2} & \textbf{70} & \textbf{68.3} & \textbf{70.1} & \textbf{70.1} & \textbf{88.9} \\ \hline
    \end{tabular}
    \caption{Results for supervised relation extraction tasks. Results on rows where the model name is marked with a~* symbol are reported as published, all other numbers have been computed by us.
    SemEval 2010 Task 8 does not establish a default split for development; for this work we use a random slice of the training set with 1,500 examples.}
    \label{tab:supervised-results}
\end{table*}

\section{Architectures for Relation Learning}
\label{sec:transformers-rel}

The primary goal of this work is to develop models that produce relation representations
directly from text. Given the strong performance of recent deep transformers trained on
variants of language modeling, we adopt \citet{devlin2018bert}'s \bert~model
as the basis for our work. In this section, we explore different 
methods of representing relations with the Transformer model.

\subsection{Relation Classification and Extraction Tasks}
\label{sec:relation-tasks}


We evaluate the different methods of representation on a suite of supervised relation extraction benchmarks. 
The relation extractions tasks we use can be broadly
categorized into two types: fully supervised relation extraction, and few-shot relation matching.

For the supervised tasks, the goal is to, given a relation statement $\br$, predict a relation type $t \in \mathcal{T}$ where $\mathcal{T}$ is a fixed dictionary of relation types and $t = 0$ typically denotes a lack of relation between the entities in the relation statement.
For this type of task we evaluate on SemEval 2010 Task~8~\cite{hendrickx2009semeval}, KBP-37~\cite{ZhangW15a} and TACRED~\cite{zhang2017position}.
More formally,

In the case of few-shot relation matching, a set of candidate relation statements are ranked, and matched, according to a query relation statement. In this task, examples in the test and development sets typically contain relation types not present in the training set.
For this type of task, we evaluate on the FewRel~\cite{han2018fewrel} dataset. 
Specifically, we are given $K$ sets of $N$ labeled relation statements
$\mathcal{S}_{k} = \{(\br_0, t_0) \dots (\br_N, t_N)\}$ where $t_i \in \{1 \dots K\}$
is the corresponding relation type. The goal is to predict the $t_q \in  \{1 \dots K\}$
for a query relation statement $\br_q$.



\subsection{Relation Representations from Deep Transformers Model}
\label{sec:transformers-results}

In all experiments in this section, we start with the \bert$_{\textsc{large}}$ model made available by \citet{devlin2018bert} and train towards task-specific losses.
Since \bert{} has not previously been applied to the problem of relation representation, we aim to answer two primary modeling questions: (1) how do we represent entities of interest in the input to \bert{}, and (2) how do we extract a fixed length representation of a relation from \bert{}'s output.
We present three options for both the input encoding, and the output relation representation. Six combinations of these are illustrated in Figure~\ref{fig:model-variants}.

\subsubsection{Entity span identification}
Recall, from Section~\ref{sec:overview}, that the relation statement $\br = (\bx, \bs_1, \bs_2)$  contains the sequence of tokens $\bx$ and the entity span identifiers $\bs_1$ and $\bs_2$.
We present three different options for getting information about the focus spans $\bs_1$ and $\bs_2$ into our \bert~encoder.

\paragraph{Standard input}
First we experiment with a \bert{} model that does not have access to any explicit identification of the entity spans $\bs_1$ and $\bs_2$. 
We refer to this choice as the \textsc{standard}~input.
This is an important reference point, since we believe that \bert{} has the ability to identify entities in $\bx$, but with the \textsc{standard}~input there is no way of knowing which two entities are in focus when $\bx$ contains more than two entity mentions. 

\paragraph{Positional embeddings}
For each of the tokens in its input, \bert~also adds a segmentation embedding, primarily used to add sentence segmentation information to the model.
To address the \textsc{standard} representation's lack of explicit entity identification, we introduce two new segmentation embeddings, one that is added to all tokens in the span $\bs_1$, while the other is added to all tokens in the span $\bs_2$. 
This approach is analogous to previous work where positional embeddings
    have been applied to relation extraction \cite{zhang2017position, DBLP:journals/corr/abs-1807-03052}.
    

\paragraph{Entity marker tokens}
Finally, we augment $\bx$ with four reserved word pieces to mark the begin and end of each entity mention in the relation statement. 
We introduce the  $[E1_{start}]$, $[E1_{end}]$, $[E2_{start}]$ and $[E2_{end}]$ and modify 
$\bx$ to give
\begin{align*}
\tilde{\bx} = & [x_0\dots [E1_{start}]~x_{i} \dots x_{j-1}~[E1_{end}] \\ 
        & \dots [E2_{start}]~x_{k}\dots x_{l-1}~[E2_{end}]\dots x_{n}]. 
\end{align*}
and we feed this token sequence into \bert~instead of $\bx$. We also update the entity indices $\tilde \bs_1 = (i + 1, j + 1)$ and $\tilde \bs_2 = (k+3, l+3)$ to account for the inserted tokens.
We refer to this representation of the input as \textsc{entity markers}.

\subsection{Fixed length relation representation}
\label{sec:relation-rep}
We now introduce three separate methods of extracting a fixed length relation representation $\bh_r$ from the \bert~encoder.
The three variants rely on extracting the last hidden layers of the transformer network, which we define as $H = [\bh_{0}, ... \bh_{n}]$ for $n = |\bx|$ (or $|\tilde{\bx}|$ if entity marker tokens are used). 

\paragraph{\cls~token} Recall from Section~\ref{sec:overview} that each $\bx$ starts with a reserved \cls~token. \bert's output state that corresponds to this token is used by \citet{devlin2018bert} as a fixed length sentence representation. We adopt the \cls~output, $\bh_0$, as our first relation representation.

\paragraph{Entity mention pooling}
We obtain $\bh_r$ by max-pooling the final hidden layers  corresponding to the word pieces in each entity mention, to get two vectors $\bh_{e_1} = \textsc{maxpool}([\bh_{i} ... \bh_{j-1}])$ and $\bh_{e_2} = \textsc{maxpool}([\bh_{k} ... \bh_{l-1}])$ representing the two entity mentions. We concatenate these two vectors to get the single representation 
$\bh_{r} = \langle \bh_{e_1} | \bh_{e_2} \rangle$ where $\langle a | b \rangle$ is the concatenation of $a$ and $b$. 
We refer to this architecture as \textsc{mention pooling}.

\paragraph{Entity start state}
Finally, we propose simply representing the relation between two entities with the concatenation of the final hidden states corresponding their respective start tokens, when \textsc{entity markers} are used. 
Recalling that \textsc{entity markers} inserts tokens in $\bx$, creating offsets in $\bs_1$ and $\bs_2$, our representation of the relation is $\br_h = \langle \bh_{i} | \bh_{j+2}\rangle$.  We refer to this output representation as
   \textsc{entity start} output. Note that this can only be applied to the \textsc{entity markers} input.

Figure~\ref{fig:model-variants} illustrates a few of the variants we evaluated in this section.
In addition to defining the model input and output architecture, we fix the training loss
used to train the models (which is illustrated in Figure~\ref{fig:loss-architecture}).
In all models, the output representation from the Transformer network is fed into
a fully connected layer that either (1) contains a linear activation, or
(2) performs layer normalization~\cite{ba2016layer} on the representation. We treat the
choice of post Transfomer layer as a hyper-parameter and use the best performing layer
type for each task.

For the supervised tasks, we introduce a new classification layer
$\mathcal{W} \in \mathcal{R}^{KxH}$ where $H$ is the size of the relation representation
and $K$ is the number of relation types. The classification loss is the standard
cross entropy of the softmax of $h_{r}W^{T}$ with respect to the true relation type.

For the few-shot task, we use the dot product between relation representation of the query
statement and each of the candidate statements as a similarity score. In this case,
we also apply a cross entropy loss of the softmax of similarity scores with respect
to the true class.

We perform task-specific fine-tuning of the \bert{} model, for all variants, with the
following set of hyper-parameters:
\begin{itemize}[noitemsep]
    \small
    \item \textbf{Transformer Architecture}: 24 layers, 1024 hidden size, 16 heads 
    \item \textbf{Weight Initialization}: BERT$_{\textsc{LARGE}}$
    \item \textbf{Post Transformer Layer}: Dense with linear activation (KBP-37 and TACRED), 
    or Layer Normalization layer (SemEval 2010 and FewRel).
    \item \textbf{Training Epochs}: 1 to 10
    \item \textbf{Learning Rate (supervised)}: 3e-5 with Adam
    \item \textbf{Batch Size (supervised)}: 64
    \item \textbf{Learning Rate (few shot)}: 1e-4 with SGD
    \item \textbf{Batch Size (few shot)}: 256
\end{itemize}


Table~\ref{tab:supervised-results} shows the results of model variants on the three
supervised relation extraction tasks and the 5-way-1-shot variant of the few-shot relation classification task. For all four tasks, the model using the
\textsc{entity markers} input representation and \textsc{entity start} output representation
achieves the best scores.

From the results, it is clear that adding  positional information in the input is critical
for the model to learn useful relation representations. Unlike previous work that have
benefited from positional embeddings~\cite{zhang2017position, DBLP:journals/corr/abs-1807-03052}, the deep Transformers benefits
the most from seeing the new entity boundary word pieces (\textsc{entity markers}). It is
also worth noting that the best variant outperforms previous published models on all
four tasks. For the remainder of the paper, we will use this architecture when further
training and evaluating our  models.
\begin{table*}[h]
    \centering
    \footnotesize
    \setlength{\tabcolsep}{3pt}
    \begin{tabular}{|m{4em}|m{45em}|} \hline
         $\br_{A}$ &
         In 1976, {\color{red} \boldmath{$e_{1}$}} (then of Bell Labs) published
         {\color{blue} \boldmath{$e_{2}$}}, the first of his books on programming
         inspired by the Unix operating system. \\ \hline
         $\br_{B}$ & 
         The ``{\color{blue} \boldmath{$e_{2}$}}'' series spread the essence of ``C/Unix
         thinking''  with makeovers for Fortran and Pascal.
         {\color{red} \boldmath{$e_{1}$}}'s Ratfor was eventually put in the public domain.\\ \hline
         $\br_{C}$ & 
         {\color{red} \boldmath{$e_{1}$}} worked at Bell Labs alongside
          {\color{olive} \boldmath{$e_{3}$}} creators Ken Thompson and Dennis Ritchie. \\ \hline
         \textbf{Mentions} &
         {\color{red} \boldmath{$e_{1}$}} = Brian Kernighan,
         {\color{blue} \boldmath{$e_{2}$}} = Software Tools,
         {\color{olive} \boldmath{$e_{3}$}} = Unix \\ \hline 
    \end{tabular}
    \caption{Example of ``matching the blanks'' automatically generated training data.
    Statement pairs $r_{A}$ and $r_{B}$ form a positive example since they share resolution
    of two entities.
    Statement pairs $r_{A}$ and $r_{C}$ as well as $r_{B}$ and $r_{C}$ form strong negative
    pairs since they share one entity in common but contain other non-matching entities.}
    \label{tab:matching-the-blanks-example}
\end{table*}

\section{Learning by Matching the Blanks}
\label{sec:mtb}

So far, we have used human labeled training data to train our relation statement encoder $\relenc$.
Inspired by open information extraction~\cite{banko2007,angeli2015leveraging}, which derives relations
directly from tagged text, we now introduce a new method of training
$\relenc$ without a predefined ontology, or relation-labeled
training data.  Instead, we declare that for any pair of relation statements $\br$ and
$\br'$, the inner product
$\relenc(\br)^{\top} \relenc(\br')$ should be high
if the two relation statements, $\br$ and $\br'$, express semantically similar relations. 
And, this inner product should be low if the two relation statements express semantically
different relations.

Unlike related work in distant supervision for information
extraction~\cite{hoffmann2011knowledge,mintz2009distant}, we do not use relation labels at training time.  Instead, we observe that there is a high degree
of redundancy in web text, and each relation between an arbitrary pair of entities is likely
to be stated multiple times. Subsequently, $\br = (\bx, \bs_1, \bs_2)$ is more likely to encode the
same semantic relation as $\br' = (\bx', \bs'_1, \bs'_2)$ if $\bs_1$ refers to the same entity as
$\bs'_1$, and $\bs_2$ refers to the same entity as $\bs'_2$.  Starting with this observation,
we introduce a new method of learning $f_{\theta}$ from entity linked text. 
We introduce this method of learning by \emph{matching the blanks} (MTB). 
In Section~\ref{sec:results} we show that MTB learns relation representations that can be used without any further tuning for relation extraction---even beating previous work that trained on human labeled data.

\subsection{Learning Setup}
\label{sec:mtb-objective}

\newcommand{\relstatement}[1]{(\bx^{#1}, \bs_1^{#1}, \bs_2^{#1})}
\newcommand{\traintuple}[1]{(\br^{#1}, e_1^{#1}, e_2^{#1})}

Let $\mathcal{E}$ be a predefined set of entities. And let $\mathcal{D} = [\traintuple{0} \dots \traintuple{N}]$ be a corpus of relation statements that have been labeled with two entities $e^i_1 \in \mathcal{E}$ and $e^i_2 \in \mathcal{E}$. 
Recall, from Section~\ref{sec:overview}, that $\br^i = (\bx^i, \bs_1^i, \bs_2^i)$, where $\bs^i_1$ and $\bs^i_2$ delimit entity mentions in $\bx^i$.
Each item in $\mathcal{D}$ is created by pairing the relation statement $\br^i$ with the two entities $e_1^i$ and $e_2^i$ corresponding to the spans $\bs_1^i$ and $\bs_2^i$, respectively.

We aim to learn a relation statement encoder $\relenc$ that we can use to determine whether or not two relation statements encode the same relation. To do this, we define the following binary classifier
\[
    p(l=1 | \br, \br') = \frac{1}{1+\exp \relenc(\br)^{\top}\relenc(\br')}
\]  
to assign a probability to the case that $\br$ and $\br'$ encode the same relation $(l=1)$, or not $(l=0)$.
We will then learn the parameterization of $\relenc$ that minimizes the loss
\begin{align}
\label{eqn:mtb-loss}
\mathcal{L(\mathcal{D})} = - &\frac{1}{|\mathcal{D}|^2} \sum_{(\br, e_1, e_2) \in \mathcal{D}} \sum_{(\br', e'_1, e'_2) \in \mathcal{D}} \\
& \delta_{e_1, e'_1}\delta _{e_2, e'_2}  \cdot \log p(l=1| \br, \br') + \nonumber \\
& (1 - \delta_{e_1, e'_1}\delta _{e_2, e'_2}) \cdot \log (1- p(l=1| \br, \br')) \nonumber
\end{align}
where $\delta_{e,e'}$ is the Kronecker delta that takes the value $1$ iff $e=e'$, and $0$ otherwise.

\subsection{Introducing Blanks}
\label{sec:introducing-blanks}
Readers may have noticed that the loss in Equation~\ref{eqn:mtb-loss} can be minimized perfectly by the entity linking system used to create $\mathcal{D}$.
And, since this linking system does not have any notion of relations, it is not reasonable to assume that $\relenc$ will somehow magically build meaningful relation representations. 
To avoid simply relearning the entity linking system, we introduce a modified corpus
\[
\tilde{\mathcal{D}} = [(\tilde{\br}^0, e_1^0, e_2^0)\dots (\tilde{\br}^N, e_1^N, e_2^N)]
\]
\noindent
where each $\tilde{\br}^i = (\tilde{\bx}^i, \bs^i_1, \bs^i_2)$ contains a relation statement in which one or both entity mentions may have been replaced by a special \blank~symbol.
Specifically, $\tilde{\bx}$ contains the span defined by $\bs_1$ with probability $\alpha$.
Otherwise, the span has been replaced with a single \blank~symbol.
The same is true for $\bs_2$.
Only $\alpha^2$ of the relation statements in $\tilde{D}$ explicitly name both of the entities that participate in the relation.
As a result, minimizing $\mathcal{L}(\tilde{\mathcal{D}})$ requires $\relenc$ to do more than simply identifying named entities in $\br$.
We hypothesize that training on $\tilde{D}$ will result in a $\relenc$ that encodes the semantic relation between the two possibly elided entity spans.
Results in Section~\ref{sec:results} support this hypothesis.

\subsection{Matching the Blanks Training}
\label{sec:mtb-implementation}


To train a model with matching the blank task, we construct a training setup similar to BERT,
where two losses are used concurrently: the masked language model loss and the matching
the blanks loss. For generating the training corpus, we use English Wikipedia and extract
text passages from the HTML paragraph blocks, ignoring lists, and tables. We use an
off-the-shelf entity linking
system\footnote{We use the public Google Cloud Natural Language API to annotate our
corpus extracting the ``entity analysis'' results --- \url{https://cloud.google.com/natural-language/docs/basics\#entity_analysis} .}
to annotate text spans with a unique knowledge base identifier (e.g.,
Freebase ID or Wikipedia URL). The span annotations include not only proper names, but
other referential entities such as common nouns and pronouns. From this annotated corpus we
extract relation statements where each statement contains
at least two grounded entities within a fixed sized window of 
tokens\footnote{We use a window of 40 tokens, which we observed provides some coverage of
long range entity relations, while avoiding a large number of co-occurring but unrelated
entities.}. To prevent a large bias towards relation statements that involve popular
entities, we limit the number of relation statements that contain the same entity by
randomly sampling a constant number of relation statements that contain any given entity.

We use these statements to train model parameters to minimize $\mathcal{L}(\tilde{\mathcal{D}})$ as described in the previous section.
In practice, it is not possible to compare every pair of relation statements, as in Equation~\ref{eqn:mtb-loss}, and so we use a noise-contrastive estimation ~\cite{gutmann2012noise, mnih2013learning}.
In this estimation, we consider all positive pairs of relation statements that contain the same entity, so there is no change to the contribution of the first term in Equation~\ref{eqn:mtb-loss}---where $\delta_{e_1,e_1'}\delta_{e_2,e_2'}=1$.
The approximation does, however, change the contribution of the second term. 

Instead of summing over all pairs of relation statements that do not contain the same pair of entities, we sample a set of negatives that are either randomly sampled uniformly from the set of all relation statement pairs, or are sampled from the set of relation statements that share just a single entity. 
We include the second set `hard' negatives to account for the fact that most randomly sampled relation statement pairs are very unlikely to be even remotely topically related, and we would like to ensure that the training procedure sees pairs of relation statements that refer to similar, but different, relations.
Finally, we probabilistically replace each entity's mention with \blank~symbols, with a probability of $\alpha=0.7$, as described in Section~\ref{sec:transformers-results}, to ensure that the model is not confounded by the absence of \blank~symbols in the evaluation tasks. In total, we generate 600 million relation statement pairs from English Wikipedia,
roughly split between 50\% positive and 50\% strong negative pairs.

\begin{figure*}[h]
  \centering
  \footnotesize
  \setlength{\tabcolsep}{3pt}
  \newcolumntype{P}[1]{>{\centering\arraybackslash}p{#1}}
  \begin{tabular}{P{0.48\textwidth} | P{0.48\textwidth}}
    \includegraphics[width=0.45\textwidth]{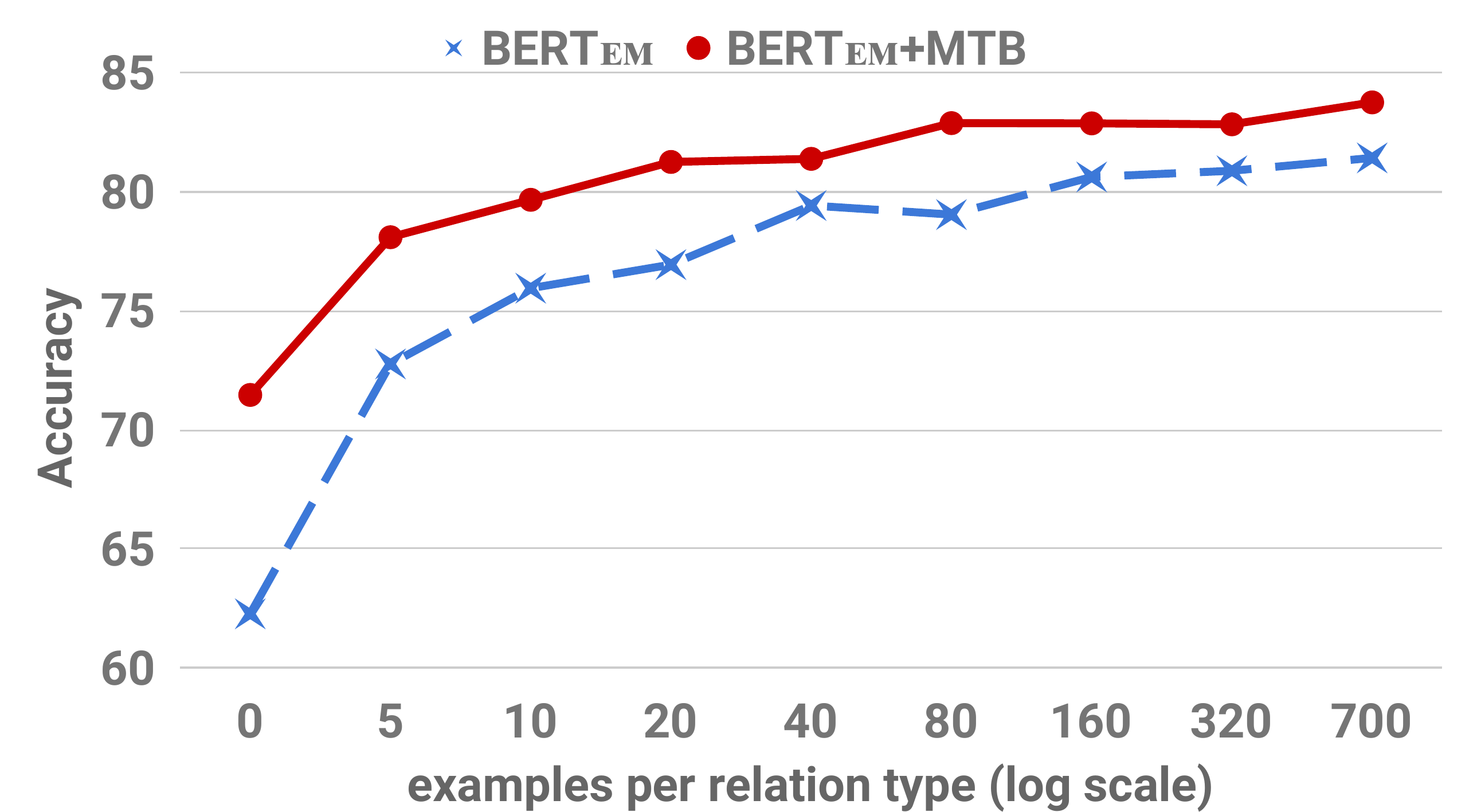} &
    \includegraphics[width=0.45\textwidth]{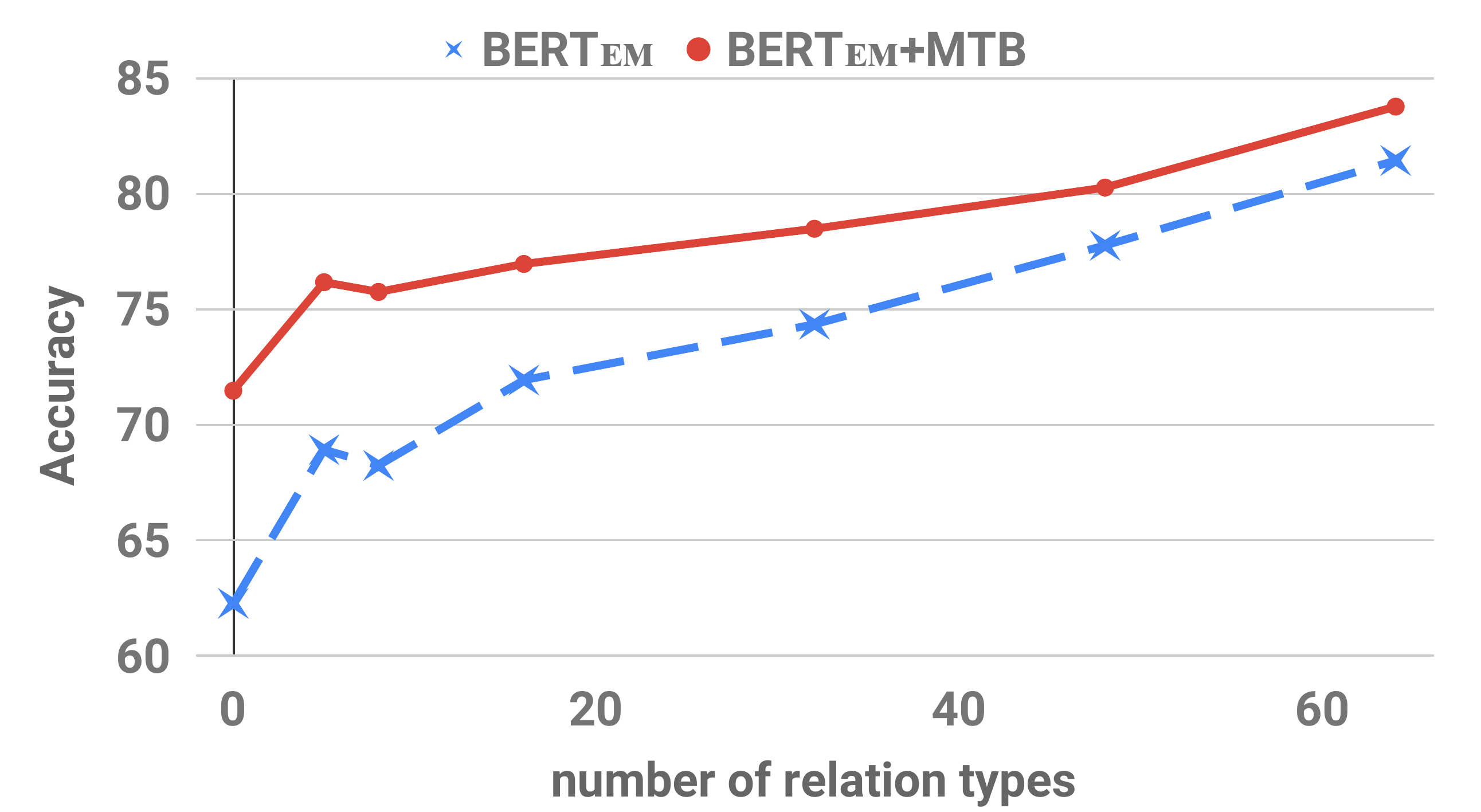} \\[8pt] 
    \begin{tabular}{|c|c|c|c|c|c|c|} \hline
      \multicolumn{7}{|c|}{\textbf{5 way 1 shot}} \\ \hline
             \# examples per type & 0    & 5    & 20   & 80   & 320  & 700 \\ \hline
      Prot.Net. (CNN)             & -- & --  & -- & -- & -- & 71.6 \\ \hline
      \bertem{}     & 72.9 & 81.6 & 85.1 & 86.9 & 88.8 & 88.9 \\ \hline
      \mtbmodel{}   & 80.4 & 85.5 & 88.4 & 89.6 & 89.6 & 90.1 \\ \hline \hline
      \multicolumn{7}{|c|}{\textbf{10 way 1 shot}} \\ \hline
              \# examples per type & 0    & 5    & 20   & 80   & 320  & 700  \\ \hline
               Prot.Net. (CNN)     & --  & --   & --   & --  &  --  & 58.8 \\ \hline
      \bertem{}     & 62.3 & 72.8 & 76.9 & 79.0 & 81.4 & 82.8 \\ \hline
      \mtbmodel{}   & 71.5 & 78.1 & 81.2 & 82.9 & 83.7 & 83.4 \\ \hline
    \end{tabular}   &
    \begin{tabular}{|c|c|c|c|c|c|} \hline
      \multicolumn{6}{|c|}{\textbf{5 way 1 shot}} \\ \hline
      \# training types  & 0     & 5     & 16    & 32    & 64 \\ \hline
      Prot.Net. (CNN)       &  -- &  --  & -- & -- & 71.6 \\ \hline
      \bertem{}     & 72.9  & 78.4  & 81.2  & 83.4  & 88.9 \\ \hline
      \mtbmodel{}   & 80.4  & 84.04 & 85.5  & 86.8  & 90.1 \\ \hline \hline
      \multicolumn{6}{|c|}{\textbf{10 way 1 shot}} \\ \hline
      \# training types & 0     & 5     & 16    & 32    & 64    \\ \hline
      Prot.Net. (CNN)       &  --  & --  & -- & -- & 58.8 \\ \hline
      \bertem{}             & 62.3  & 68.9  & 71.9  & 74.3  & 81.4 \\ \hline
      \mtbmodel{}           & 71.5 & 76.2 & 76.9 & 78.5 & 83.7  \\ \hline
    \end{tabular}   \\ 
  \end{tabular}
  \caption{Comparison of classifiers tuned on FewRel. Results are for the development set
           while varying
           the amount of annotated examples available for fine-tuning. On the left,
           we display accuracies while
           varying the number of examples per relation type, while maintaining all 64 relations
           available for training. On the right, we display accuracy on the development set
           of the two models while varying the total number of relation types available for
           tuning, while maintaining all 700 examples per relation type. In both graphs,
           results for the 10-way-1-shot variant of the task are displayed.}
  \label{fig:fewrel-limit-examples}
\end{figure*}


\section{Experimental Evaluation}
\label{sec:results}

In this section, we evaluate the impact of training by matching the blanks. 
We start with the best \bert{} based model from Section~\ref{sec:relation-rep}, which we call \bertem{}, and we compare this to a variant that is trained with the matching the blanks task (\mtbmodel{}).
We train the \mtbmodel{} model by initializing the Transformer weights to
the weights from \bert$_{\textsc{LARGE}}$ and use the following parameters:
\begin{itemize}[noitemsep]
    \small
    \item \textbf{Learning rate}: 3e-5 with Adam
    \item \textbf{Batch size}: 2,048
    \item \textbf{Number of steps}: 1 million
    \item \textbf{Relation representation}: \textsc{Entity Marker}
\end{itemize}

We report results on all of the tasks from Section~\ref{sec:relation-tasks}, using the same
task-specific training methodology for both \bertem{} and \mtbmodel{}.



\subsection{Few-shot Relation Matching}

\begin{table}[]
    \centering
    \footnotesize
    \begin{tabular}{|c|c|c|c|c|} \hline
    & \textbf{5-way} & \textbf{5-way} & \textbf{10-way} & \textbf{10-way} \\
    & \textbf{1-shot} & \textbf{5-shot} & \textbf{1-shot} & \textbf{5-shot} \\
    \hline
Proto Net  & 69.2  & 84.79  & 56.44 & 75.55 \\ \hline 
\mtbmodel{}                 &  \textbf{93.9} & \textbf{97.1} & \textbf{89.2} & \textbf{94.3} \\ \hline 
Human                       & 92.22 &  --   & 85.88  & -- \\ \hline
    \end{tabular}
    \caption{Test results for FewRel few-shot relation classification task. Proto Net is the best published system from \citet{han2018fewrel}.  At the time of writing, our \mtbmodel{} model outperforms the top model on
    the leaderboard (\url{http://www.zhuhao.me/fewrel/}) by over 10\% on the 5-way-1-shot and over 15\% on the 10-way-1-shot configurations.}
    \label{tab:fewshot-results}
\end{table}

First, we investigate the ability of \mtbmodel{} to solve the FewRel task without any
task-specific training data. 
Since FewRel is an exemplar-based approach, we can just rank each candidate relation statement according to its representation's inner product with the exemplars' representations.

Figure~\ref{fig:fewrel-limit-examples} shows that the task agnostic \bertem{} and \mtbmodel{} models outperform the previous published state of the art on FewRel task even when they have not seen any FewRel training data.
For \mtbmodel{}, the increase over \citet{han2018fewrel}'s supervised approach is very significant---8.8\% on the 5-way-1-shot task and 12.7\% on the 10-way-1-shot task. 
\mtbmodel{} also significantly outperforms \bertem{} in this unsupervised setting, which is to be expected since there is no relation-specific loss during \bertem{}'s training. 

To investigate the impact of supervision on \bertem{} and \mtbmodel{}, we introduce increasing amounts of FewRel's training data. Figure~\ref{fig:fewrel-limit-examples} shows the increase in performance as we either increase the number of training examples for each relation type, or we increase the number of relation types in the training data.
When given access to all of the training data, \bertem{} approaches \mtbmodel{}'s performance. However, when we keep all relation types during training, and vary the number of types per example, \mtbmodel{} only needs 6\% of the training data to match the performance of a \bertem{} model trained on all of the training data.
We observe that maintaining a diversity of relation types, and reducing the number of examples per type, is the most effective way to reduce annotation effort for this task. 
The results in Figure~\ref{fig:fewrel-limit-examples} show that MTB training could be used to significantly reduce effort in implementing an exemplar based relation extraction system.

Finally, we report \mtbmodel{}'s performance on all of FewRel's fully supervised tasks in Table~\ref{tab:fewshot-results}.
We see that it outperforms the human upper bound reported by \citet{han2018fewrel}, and it significantly outperforms all other submissions to the FewRel leaderboard, published or unpublished.

\subsection{Supervised Relation Extraction}


\begin{table}[]
    \centering
    \setlength{\tabcolsep}{3pt}
    \footnotesize
    \begin{tabular}{|c|c|c|c|} \hline
                     & \textbf{SemEval 2010} & \textbf{KBP37} & \textbf{TACRED} \\ \hline
SOTA        & 84.8                  & 58.8           & 68.2         \\ \hline
\bertem{}   & 89.2                  &   68.3         & 70.1            \\ \hline
\mtbmodel{} &\textbf{89.5}          &  \textbf{69.3} & \textbf{71.5}   \\ \hline
    \end{tabular}
    \caption{F1 scores of \mtbmodel{} and \bertem{} based relation classifiers on the
    respective test sets. Details of the SOTA systems are given in Table~\ref{tab:supervised-results}.}
    \label{tab:mtb-classifiers}
\end{table}

\begin{table}[]
    \centering
    \setlength{\tabcolsep}{3pt}
    \footnotesize
    \begin{tabular}{|c|c|c|c|c|c|} \hline
    \textbf{\% of training set} &  \textbf{1\%} & \textbf{10\%} & \textbf{20\%} & \textbf{50\%} & \textbf{100\%} \\ \hline \hline
     \textbf{SemEval 2010 Task 8} & \multicolumn{5}{c|}{}  \\ \hline
     \bertem{}       & 28.6 & 66.9 & 75.5 & 80.3 & 82.1 \\ \hline
     \mtbmodel{}     & 31.2 & 70.8 & 76.2 & 80.4 & 82.7 \\ \hline \hline
     \textbf{KBP-37}              &  \multicolumn{5}{c|}{}  \\ \hline
     \bertem{}       & 40.1 & 63.6 & 65.4 & 67.8 & 69.5 \\ \hline
     \mtbmodel{}     & 44.2 & 66.3 & 67.2 & 68.8 & 70.3  \\ \hline \hline
     \textbf{TACRED }             &  \multicolumn{5}{c|}{}   \\ \hline     
     \bertem{}       & 32.8 & 59.6 & 65.6 & 69.0 & 70.1 \\ \hline
     \mtbmodel{}     & 43.4 & 64.8 & 67.2 & 69.9 & 70.6 \\ \hline
     \end{tabular}
    \caption{F1 scores on development sets for supervised relation extraction
    tasks while varying the amount of tuning data available to our \bertem{} and \mtbmodel{} models.}
    \label{tab:supervised-limit-train}
\end{table}

Table~\ref{tab:mtb-classifiers} contains results for our classifiers tuned on supervised relation extraction data. As was established in Section~\ref{sec:transformers-results}, our \bertem{} based classifiers outperform previously
published results for these three tasks. The additional MTB based training further increases
F1 scores for all tasks.


We also analyzed the performance of our two models while reducing the amount of supervised
task specific tuning data. The results displayed in Table~\ref{tab:supervised-limit-train}
show the development set performance when tuning on a random subset of the task specific
training data. For all tasks, we see that MTB based training is even more effective for 
low-resource cases, where there is a larger gap in performance between our \bertem{} and
\mtbmodel{} based classifiers.
This further supports our argument that training by matching the blanks can significantly reduce the amount of human input required to create relation extractors, and populate a knowledge base.





\section{Conclusion and Future Work}

In this paper we study the problem of producing useful relation representations directly
from text. We describe a novel training setup, which we call \textit{matching the blanks},
which relies solely on entity resolution annotations. When coupled with a new architecture for 
fine-tuning relation representations in BERT, our models achieves state-of-the-art results on
three  relation extraction tasks, and outperforms human accuracy on few-shot relation matching.
In addition, we show how the new model is particularly effective
in low-resource regimes, and we argue that it could significantly reduce the amount of human effort required to create relation extractors.

In future work, we plan to work on \emph{relation discovery} by clustering relation 
statements that have similar representations according to \mtbmodel.
This would take us some of the way toward our goal of truly general purpose relation identification and extraction.
We will also study representations of relations and entities that can be used to store relation triples in a distributed knowledge base. 
This is inspired by recent work in knowledge base embedding \cite{bordes2013translating,nickel2016holographic}.

\newpage
\bibliography{acl2019}
\bibliographystyle{acl_natbib}


\end{document}